\documentclass[10pt,twocolumn,letterpaper]{article}
\usepackage{wacv}

\usepackage{xcolor} 
\usepackage[table]{xcolor}

\definecolor{linkblue}{rgb}{0.21,0.49,0.74}
\usepackage[pagebackref,breaklinks,colorlinks,allcolors=linkblue]{hyperref}
\usepackage{amssymb} 
\usepackage[normalem]{ulem} 
\usepackage{amsmath} 
\usepackage{pifont}
\usepackage{bm}

\usepackage{booktabs}
\usepackage{arydshln}
\definecolor{lightgray}{gray}{0.96}
\newcommand{\cmark}{\ding{51}} 
\newcommand{\xmark}{\ding{55}} 
\usepackage{multirow} 
\usepackage{makecell} 
\usepackage{placeins}
\newcommand{\pmt}{\!\pm\!}
\DeclareMathOperator*{\argmax}{arg\,max}
\usepackage{algorithm}
\usepackage{algpseudocode}
\usepackage{pgfplots}

\title{SpurCon: Weighted Supervised Contrastive Learning for Mitigating Spurious Cues in Medical Imaging}
\author{Shenhav Nadir\textsuperscript{*} \quad
Meir Yossef Levi \quad
Eyal Gofer\textsuperscript{\dag} \quad
Guy Gilboa\textsuperscript{\dag}\\[1ex]
Viterbi Faculty of Electrical and Computer Engineering, Technion - Israel Institute of Technology\\
{\small\textsuperscript{*}\texttt{shenhav.n@campus.technion.ac.il}
}
}
\makeatletter
\let\savedmaketitle\maketitle
\let\savedinternalmaketitle\@maketitle
\makeatother

\begin{document}
\AddToHookNext{shipout/foreground}{%
  \put(54,-758){%
    \makebox[0pt][l]{%
      \scriptsize\textsuperscript{\dag} Joint supervision.
    }%
  }%
}
\maketitle

\begin{abstract}
Despite the rapid progress of deep neural networks in visual recognition, their adoption in high-risk medical applications remains limited due to reliability and robustness concerns. Models may exploit spurious correlations, particularly in medical imaging, where devices or treatment artifacts often co-occur with pathology. In small or imbalanced datasets, such cues further reduce worst-group performance and undermine clinical trust.
To solve these issues, two major challenges should be addressed: identifying dataset-specific spurious cues, which typically require domain knowledge, and mitigating reliance on them. To tackle both, we propose \textit{\textbf{SpurCon}}, a lightweight framework based on a novel supervised contrastive loss formulation that leverages available metadata and predicted spurious labels to enhance robustness.
We introduce a fast few-shot procedure, without network training, to estimate spurious labels using a small number of expert-annotated samples. We then propose a weighted supervised contrastive objective, \textit{WtSupCon}, that reshapes the representation geometry by assigning sample-specific weights that depend on the [pathology, spurious, metadata] combination. For example, the highest weight is assigned to samples that differ only in their spurious label. This yields highly similar representations for images with the same metadata and pathology, differing only in the predicted spurious label. Our method operates on pretrained image encoders (such as BiomedCLIP) and trains only a lightweight projection head.
We evaluate \textit{\textbf{SpurCon}} on a synthetic setting and on Waterbirds, CheXpert, a chest X-ray classification dataset, and ISIC 2020, a skin cancer classification dataset. Our approach delivers the best spurious-mitigation performance, balancing well worst-group and overall accuracy on multiple datasets. 
\end{abstract}
    
\section{Introduction}
\label{sec:intro}
 \begin{figure*}
   \centering
\includegraphics[width=0.8\linewidth]{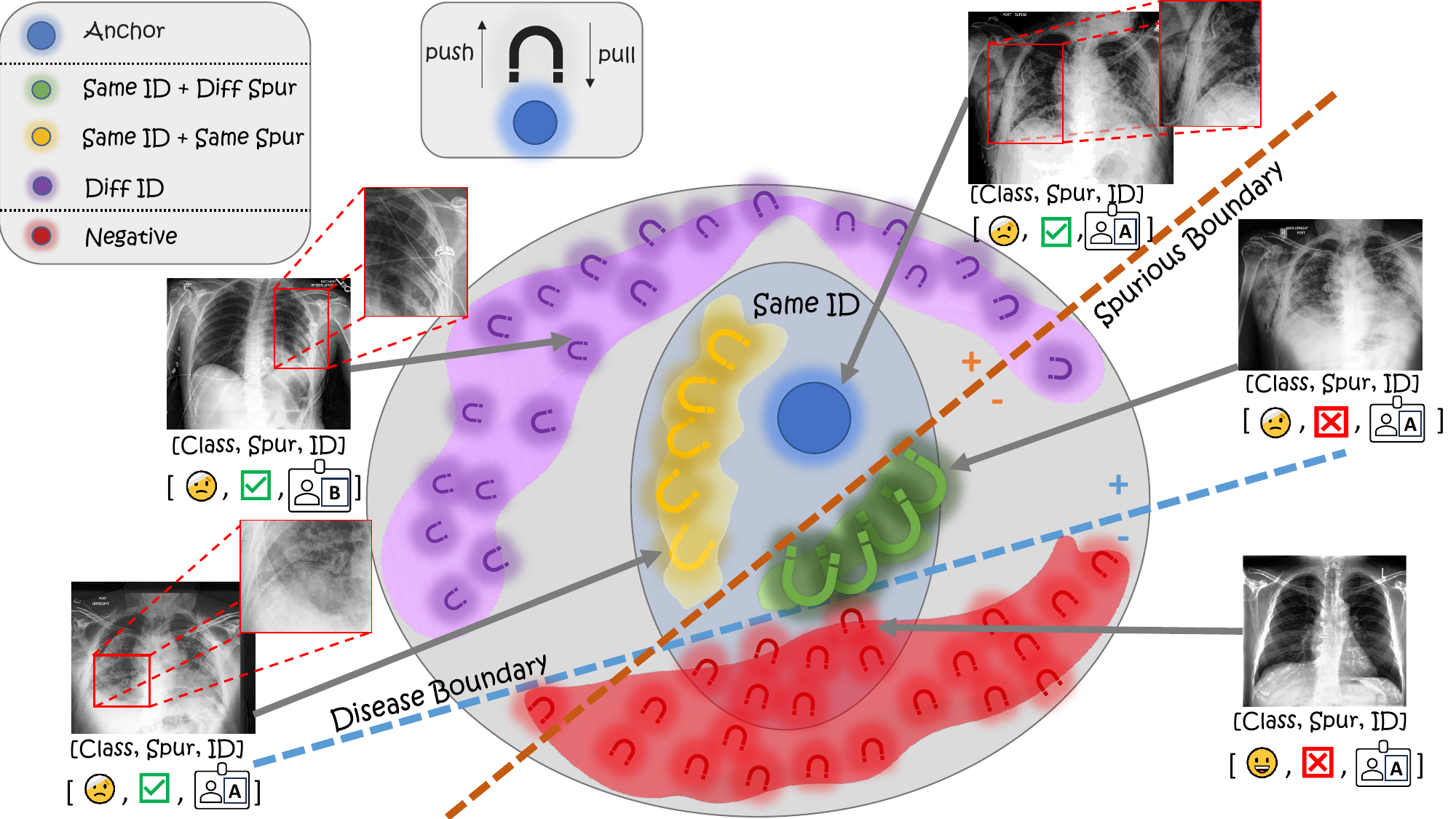}
\caption{\textbf{SpurCon overview.} Given an anchor, we define anchor-sample pairs using [class, spurious, ID]. Pairs with different classes (\eg, pathology) are pushed apart (negatives). Positive pairs are pulled together, with decreasing strength: (1) same ID \& different spurious cue, (2) same ID \& same spurious cue, (3) same class only. } \label{fig:teaser}
\end{figure*}
Deep learning can substantially improve medical imaging diagnosis by learning expressive visual representations that help streamline clinical workflows~\cite{huang2023self,waller2022applications}. However, the adoption of deep learning-based algorithms in high-risk medical applications remains limited due to concerns about reliability and robustness \cite{lekadir2025future,vollmer2020machine}. In particular, learned representations may capture spurious attributes, i.e., input cues that are predictive in-distribution but are not necessarily target-related \cite{ye2024spurious,saab2022reducing,sagawa2019distributionally}. 
While spurious correlations arise across many domains, including classical computer vision~\cite{sagawa2019distributionally}, natural language processing~\cite{du2023shortcut}, and fairness-sensitive applications~\cite{obermeyer2019dissecting}, they are particularly problematic in medical imaging. Medical images often contain non-pathological cues (\eg, tubes or surgical markings) that systematically co-occur with specific pathologies, encouraging models to rely on shortcut cues rather than clinically meaningful features~\cite{saab2022reducing} (see~\cref{fig:spur_examples}).

Addressing spurious correlations can be broadly decomposed into two sub-problems: \textit{identifying} the spurious attributes and \textit{mitigating} their effects. For the former, identifying which attributes should be considered spurious is often non-trivial, especially in the medical domain, where domain expertise is required to determine which cues are undesirable for the decision-making process of the model \cite{ye2024spurious}. Fully manual annotation of spurious attributes is impractical at scale. Therefore, prior approaches \cite{sohoni2021barack,nam2022spread} rely on auxiliary or ad hoc classifiers for automatic labeling based on a small set of spurious-labeled samples. However, these methods fail to exploit the rich semantic knowledge embedded in modern foundation models.
Regarding mitigation, one promising direction is to improve geometric representation alignment via contrastive learning. Prior studies \cite{zhang2022contrastive,zhang2022correct} introduce supervised contrastive approaches that align representations of same-class samples across different spurious annotations, inferred by unsupervised methods. However, these works were developed and evaluated primarily on natural-image benchmarks, where the spurious attributes differ from those in medical imaging, resulting in poor performance on medical imaging datasets (For comparison with multiple prior methods, see~\cref{tab:spurcon-results}). We conclude that class and spurious-attribute information alone are insufficient for spurious mitigation in the medical domain, and that incorporating additional clinical context is essential. 

We propose \textbf{\textit{SpurCon}}, a lightweight framework built on a novel supervised contrastive loss formulation that leverages available metadata and predicted spurious labels to enhance robustness. First, we propose a few-shot spurious-label inference method based on expert-selected examples that evenly cover the (label, spurious) combinations. Using a powerful pretrained image encoder, we extract meaningful visual embeddings for each spurious attribute, compute their mean, and assign labels to the remaining samples by nearest neighbor in terms of cosine similarity.
Then, inspired by prior work~\cite{vu2021medaug}, we propose a weighted supervised contrastive loss, \textit{\textbf{WtSupCon}}. It assigns sample-specific weights that depend on the [pathology, spurious, metadata] combination, with the highest weight given to pairs that differ only in their spurious label (see Fig. \ref{fig:teaser}). Built on strong pretrained image representations, our method trains only a lightweight projection head, improving runtime compared to prior strong spurious-mitigation studies. 

While our approach is well suited to medical imaging datasets, it is not limited to them. We evaluate \textit{SpurCon} on a synthetic setting and on three publicly available datasets: Waterbirds~\cite{sagawa2019distributionally}, CheXpert~\cite{irvin2019chexpert}, a chest X-ray classification dataset and ISIC 2020~\cite{rotemberg2021patient}, a skin cancer classification dataset.
Our main contributions can be summarized as follows: 
\begin{itemize}
\renewcommand{\labelitemi}{$\bullet$}
    \item We propose a simple yet reliable few-shot approach for estimating spurious labels using powerful pretrained image encoders. 
  \item Building on these estimated labels, we introduce SpurCon, a lightweight method based on a novel supervised contrastive loss, WtSupCon, that leverages available metadata (\eg, patient ID) to reduce reliance on spurious attributes.
  \item SpurCon achieves the strongest spurious-mitigation performance across both medical and non-medical datasets, regularly improving average and worst-group accuracies while substantially reducing runtime. 
\end{itemize}

\section{Method}
\label{sec:method}
\textbf{Preliminaries.} Let $\mathcal{D} = \{(x_j, y_j, s_j)\}_{j=1}^{M}$ denote a dataset of triplets, where $x_j$ is an image, $y_j \in \mathcal{Y}=\{1,\dots,C\}$ is its class label, and $s_j \in \mathcal{S}=\{1,\dots,S\}$ is the corresponding spurious attribute. We define the group assignment as $g_j = (y_j, s_j) \in \mathcal{G} = \mathcal{Y} \times \mathcal{S}$. The embedding of image $x_j$ by the powerful image encoder is denoted by $\textbf{z}_j\in\mathbb{R}^d$.
We denote $\langle a,b \rangle$ for the cosine similarity $\frac{a^\top b}{\|a\|\|b\|}$.

\subsection{Pick-and-Predict: few-shot learning for predicting spurious labels} \label{pick-and-predict}
Prior methods \cite{nam2022spread,sohoni2021barack} that use a fixed number of spurious-labeled samples train a dedicated model to predict spurious labels. In contrast, we infer spurious labels from a small set of expert-selected examples via a few-shot procedure, without any additional model training.
For each spurious label $s \in \{1,\dots,S\}$, we define a spurious-label prototype as the prototype of the $N$ expert-annotated samples assigned to that label: $\boldsymbol{\mu}^{(s)} = \frac{1}{N} \sum_{i=1}^{N} \mathbf{z}_i^{(s)},$ where  $\{\mathbf{z}_i^{(s)}\}_{i=1}^{N}$ are the embeddings of the $N$ expert-annotated samples per label, $\{x_i^{(s)}\}_{i=1}^{N}$, evenly covering $\mathcal{G}$. 
Given a new sample $x_j$, its inferred spurious label is assigned by
\begin{equation}
\hat{s}_j=\argmax_{s\in\{1,\ldots,S\}}
\left(\langle \mathbf{z}_j, \boldsymbol{\mu}^{(s)} \rangle \right).
\label{eq:cos-sim}
\end{equation}

\begin{figure*}[t]
    \centering
    \includegraphics[width=\textwidth]{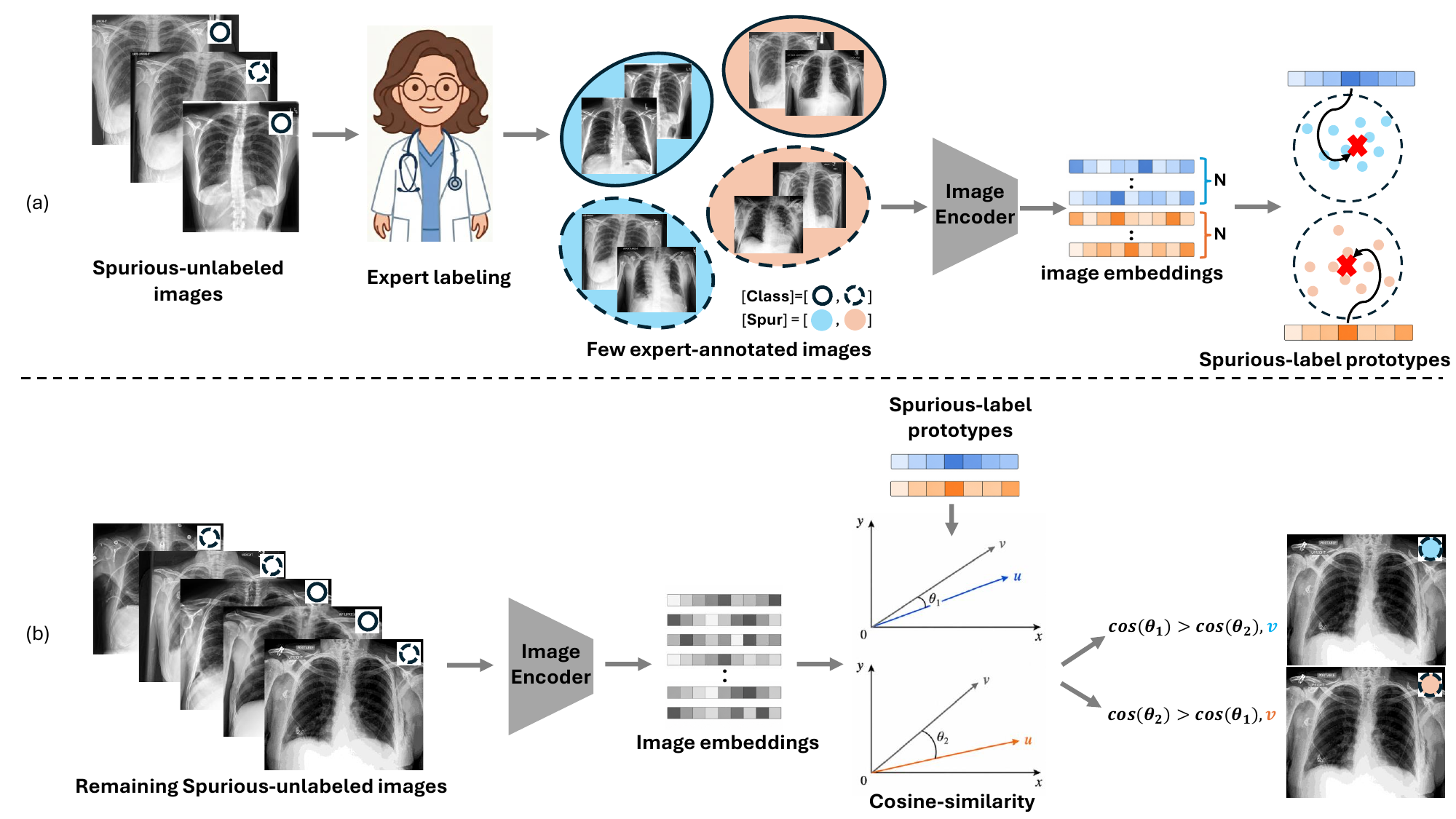}
    \caption[Pick-and-Predict method overview]{\textbf{Pick-and-Predict overview.} Illustration of \textbf{reliable} and \textbf{annotation-efficient} few-shot spurious-label estimation, shown for binary class and spurious attributes. (a) Given known training class labels, an expert annotates $N$ samples per spurious category, evenly distributed across classes. Their embeddings are averaged to form a prototype for each spurious label. (b) Each remaining training or validation sample is encoded and assigned the spurious label whose prototype has the highest cosine similarity.}
    \label{fig:pick_n_predict_overview}
\end{figure*}

\noindent\Cref{fig:pick_n_predict_overview} illustrates this method. 

\subsection{Weighted-SupCon: reducing spurious attribute effects in foundation-model representations} \label{weighted_supcon}
Supervised contrastive learning~\cite{khosla2020supervised} leverages class-label information by pulling together representations of samples that share the same label and pushing apart samples with different labels. The supervised contrastive (SupCon) loss $\mathcal{L}^{\mathrm{sc}}$ is defined as 

\begin{equation}
\label{supcon}
\mathcal{L}^{\mathrm{sc}}
= \sum_{i \in I} \mathcal{L}^{\mathrm{sc}}_i
=
\sum_{i \in I}
\left[-
\frac{1}{|P(i)|}
\sum_{p \in P(i)}
\ell_{i,p}\right],
\end{equation}
where
\begin{equation}
\label{supcon_ellip}
\ell_{i,p}
=
\log
\frac{
\exp\!\left(\langle \mathbf{z}_i, \mathbf{z}_p \rangle / \tau\right)
}{
\sum_{a \in A(i)}
\exp\!\left(\langle \mathbf{z}_i, \mathbf{z}_a \rangle / \tau\right)
}.
\end{equation}

Here, $I$ is the index set of all samples in a batch; $A(i)=I\setminus \{i\}$ is the set of all candidate indices excluding the anchor; $P(i)=\{p \in A(i)\,:\, y_p = y_i\}$ is the set of indices of all positives; $\tau>0$ is a temperature hyperparameter.

In Eq.~\eqref{supcon}, all positive samples contribute equally to the loss. 
However, the positive set may contain heterogeneous subsets whose influence on the representation should be controlled differently. 
We therefore introduce \textit{Weighted-SupCon loss (WtSupCon)}, a generalized formulation that partitions all positive samples into $K$ disjoint sets $\{P_k(i)\}_{k=1}^{K}$ and assigns each set a weight $\alpha_k > 0$. The resulting loss is
\begin{equation} 
\label{eq:wtsupcon}
\begin{aligned}
\mathcal{L}^{\mathrm{wsc}}
=
\sum_{i \in I}
\left[
\sum_{k=1}^{K}
-\frac{\alpha_k}{|P_k(i)|}
\sum_{p \in P_k(i)}
\ell_{i,p}
\right].
\end{aligned}
\end{equation}

\noindent \Cref{alg:method} summarizes the procedure of SpurCon.

\noindent\textbf{Split into sets.} In our context, the primary motivation for this partitioning is to mitigate spurious correlations. 
Accordingly, we split the positive set into $K=3$ sets, where the highest weight is assigned to pairs of samples that differ only in their spurious cues, whereas a milder weight is assigned to pairs that are identical in all attributes, as they contribute less to suppressing spurious cues while remaining useful for similarity. The lowest weight is assigned to pairs that match only in their class label. The weights also compensate for set imbalance, with relative weights scaled inversely with set size, so that smaller sets receive larger weights (see the number and sizes of magnets per set in~\cref{fig:teaser}). 
By emphasizing cross-spurious pairs within the same class and ID, the loss explicitly encourages representations of samples that vary in spurious attributes to reside close to one another, thereby reducing the reliance of the model on those attributes.
A complete description of the partition is provided in~\cref{weight-sets}.

\noindent\textbf{Customized sampling.} 
Our WtSupCon loss is most effective when a mini-batch contains samples from each positive-pair set. Given the definition of sets in~\cref{weight-sets}, batching together samples that share the same ID increases the likelihood that sets~1 and~2 co-occur within a mini-batch. Accordingly, we construct mini-batches that favor placing same-ID samples in the same batch, which we call the \textit{ID-paired} sampler. However, many medical datasets are highly imbalanced. In such cases, we instead aim to include samples from multiple groups $\mathcal{G}$ in every mini-batch to avoid under-representing minority groups. 
This alternative sampler prioritizes group coverage within each mini-batch rather than set coverage; we refer to this as the \textit{Balanced-groups} sampler.

\noindent\textbf{Projection head.}
The image embedding $\mathbf{z}_j$ is further fed into a lightweight projection head.
This head is instantiated in one of three configurations with tunable hidden width $h$ and dropout rate $dp$: (1) a 3-layer LayerNorm–GeLU MLP with a residual skip, (2) a 3-layer BatchNorm–ReLU MLP with a bottleneck ($h/2$), and (3) a shallower variant of (2) without the final layer. Configuration details per dataset are given in~\cref{experimental_details}.
During training, only the projection head parameters are optimized, while the image encoder remains frozen.

\noindent\textbf{Metadata usage.} We leverage metadata associated with medical images to find attributes that can serve as grouping keys for visually similar samples, such as patient ID. When metadata is unavailable, comparable grouping cues can often be inferred from dataset directory structure, as done in Waterbirds (see~\cref{experimental_details}). If no informative knowledge is available, the loss remains well-defined and effective. It reduces to a two-set formulation based only on class and spurious labels, analogous to SupCon becoming self-supervised when class labels are absent. 

\begin{table}
\centering
\setlength{\tabcolsep}{10pt}
\begin{tabular}{c|c c c}
\toprule
\makecell{Set} &  \makecell{Same \\ID} & \makecell{Same \\spurious label} & \makecell{Same \\class}\\
\midrule
1 &   \textcolor{green}{\cmark} &\textcolor{red}{\xmark} &  \textcolor{green}{\cmark}\\
2 &    \textcolor{green}{\cmark} &  \textcolor{green}{\cmark} &  \textcolor{green}{\cmark}\\
3 &  \textcolor{red}{\xmark} &{-} &  \textcolor{green}{\cmark}\\
\bottomrule
\end{tabular}

\caption{\textbf{Split into Sets.} Set definitions ordered by importance in the final loss. The highest-weight set shares ID and class with the anchor but differs in spurious label.}
\label{weight-sets}
\end{table}

\begin{algorithm}
\small
\caption{SpurCon: Spurious Correlation Mitigation}
\label{alg:method}
\begin{algorithmic}[1]
\Require Frozen encoder $E$, Projection head $H$, small expert-labeled set $Ex=\{x_i\}_{i=1}^{S\cdot N}$, Training set $Tr=\{x_j,y_j,m_j\}_{j=1}^{T}$, validation set $Val=\{x_j,y_j,m_j\}_{j=1}^{V}$, test set $Te$, $K$ weight values, classifier $log\_reg$.
\Statex
\Statex \textbf{Few-shot prediction}
\State Encode $Ex$: $\{\textbf{z}_i\}_{i=1}^{S\cdot N}$.
\State Compute prototypes:  $\boldsymbol{\mu}^{(s)}, s \in \{1,\dots,S\}$.
\State Encode $Tr$ and $Val$: $\{\mathbf{z}_j^{D'}\}_{j=1}^{|D'|}$, for $D'\in\{Tr,Val\}$.
\State Calculate predicted spurious labels: $\{\hat{s}_j^{D'}\}_{j=1}^{|D'|}$ \hfill (\text{\Cref{eq:cos-sim}}).

\Statex 
\setcounter{ALG@line}{0}
\Statex \textbf{Weighted-SupCon (train-time)}
\State Combine $Tr$ and $Val$: 
$\mathcal{D}=\{(\mathbf{z}_j,y_j,\hat{s}_j,m_j)\}_{j=1}^{T'}$

\State Split $\mathcal{D}$ into $F$ folds using StratifiedGroupKFold.

\State \textbf{for} fold $f=\{1,..,F\}$ \textbf{do}:
\State \quad Split $\mathcal{D}$ into $\mathcal{D}_{\mathrm{train}}^{(f)}$ and
$\mathcal{D}_{\mathrm{val}}^{(f)}$.
\State \quad Construct customized sampler $\mathcal{S}^{(f)}$ for
$\mathcal{D}_{\mathrm{train}}^{(f)}$.

\State \quad \textbf{for} epoch $e=1,\ldots,E_{\mathrm{train}}$ \textbf{do}:
\State \quad\quad Form batches $\{\mathbf{z}_b,\mathbf{y}_b,\hat{\mathbf{s}}_b,\mathbf{m}_b\}_{b=1}^{B}$ using
$\mathcal{S}^{(f)}$.
\State \quad\quad Construct mask per batch: $M_b(\mathbf{y}_b,\hat{\mathbf{s}}_b,\mathbf{m}_b,\{\alpha_k\}_{k=1}^{K})$.
\State \quad\quad Optimize
$WtSupCon(H(\mathbf{z}_b),M_b,\tau)$ \hfill (\text{\Cref{eq:wtsupcon}}).

\State \quad\quad Fit $\mathrm{log\_reg}\left(H(\mathbf{z}_{train}^{(f)})\right)$,  Predict $\mathrm{log\_reg}\left(H(\mathbf{z}_{val}^{(f)})\right)$.
\State \quad \textbf{end for}
\State \quad $e_f^\star \leftarrow$ epoch with the highest $val$ WG accuracy.
\State \textbf{end for}
\State $\tilde{e}\leftarrow\mathrm{round}\!\left(
\mathrm{median}\!\left(\{e_f^\star\}_{f=1}^{F}\right)\right)$.
\State Set the refit epoch:
$e^\star\leftarrow\min\!\left(E_{\mathrm{train}},\max(1,\tilde{e})\right)$.
\State $\hat{H} \leftarrow$ Train $H$ on $\mathcal{D}$ for $e^\star$ epochs.

\Statex 
\setcounter{ALG@line}{0}
\Statex \textbf{Weighted-SupCon (test-time)}
\State Load $\hat{H}$.
\State Project embeddings:
$\hat{H}(\mathbf{z}^{\mathcal{D}})$ and 
$\hat{H}(\mathbf{z}^{Te})$.
\State Fit $log\_reg\left(\hat{H}(\mathbf{z}^{\mathcal{D}})\right)$.
\State Predict and evaluate on $\hat{H}(\mathbf{z}^{Te})$.

\end{algorithmic}
\end{algorithm}
\section{Experiments and Results}
\label{sec:experiments_and_results}

\subsection{Experimental Setup} \label{experimental_details}
First, we describe our toy dataset and the publicly available datasets on which we evaluate. We then describe the implementation details used in our experiments. For all datasets, training, validation and test sets are ID-disjoint.

\noindent\textbf{Toy dataset}: a binary classification dataset of generated images with classes $\mathcal{Y}=\{one \:hole, two\:holes \}$ and two background types used as a spurious attribute:  $S=\{stripes,dots\}$, both encoded as $\{0,1\}$, respectively.
Variations include colors (=IDs), scale, jitter, blurring and random small background objects. 
The training/validation/test splits comprise 7,000/1,000/2,000 samples; the training set includes (class, spurious) combinations of $(0,0)=47.5\%$, $(0,1)=2.5\%$, $(1,0)=2.5\%$, and $(1,1)=47.5\%$, the validation set uses intermediate proportions, and the test set is balanced. See example in \cref{fig:spur_examples}, and several more in the Supplementary Material (Supp.).

\noindent\textbf{Waterbirds}~\cite{sagawa2019distributionally}: a well-known dataset for spurious evaluation with water and land birds and mixed backgrounds of water and land. Bird species, parsed from the image paths, are used as IDs.

\noindent\textbf{CheXpert - Pneumothorax}~\cite{irvin2019chexpert}: a large chest X-ray classification dataset, including patient ID and a \textit{Support Devices} column indicating the absence or presence of any medical device in the scan, which we treat as a spurious attribute. We define ``No Finding'' as the negative class. Within each class, the training set is highly imbalanced between the two groups defined by the predicted spurious labels ($\approx98\%/2\%$), whereas the test set is balanced within each class. In total, 36,679 X-ray scans were used.

\noindent\textbf{ISIC 2020}~\cite{rotemberg2021patient}: the dataset has a binary label for malignancy. We combine metadata from~\cite{bevan2021skin,rotemberg2021patient} to obtain a binary spurious attribute for the presence of rulers in the image and patient ID. Following~\cite{bevan2021skin}, we use images which are center-cropped and resized to $256\times256$, for a total of 32,692 images.
See~\cref{fig:spur_examples} for spurious examples from each dataset.

\noindent\textbf{Metrics.} 
We report average accuracy under the test distribution (\textit{Avg.}), adjusted average accuracy reweighted by true training-group proportions~\cite{sagawa2019distributionally} (\textit{Adj. Avg.}), worst-group accuracy (\textit{WG})~\cite{zhang2022correct}, and Area Under the ROC Curve (\textit{AUC}). As the primary metric for spurious correlation mitigation, {WG} is highlighted in yellow across all tables. High {Avg.} performance can easily mask a model's reliance on spurious features; for instance, the Waterbirds baseline (\cref{tab:spurcon-results}) yields strong {Avg.} accuracy but severely degraded {WG} performance. Formally, given groups $\mathcal{G}$, {WG} accuracy is the minimum performance across all $g \in \mathcal{G}$, whereas {Avg.} measures overall sample-wise correctness. Because optimizing for sub-population robustness often compromises average performance, we report both metrics, prioritizing {WG} while striving for a balanced trade-off.

\noindent\textbf{Implementation details.} \label{subsec:implementation_details} We extract image representations from pretrained image encoders: CLIP~\cite{radford2021learning} for natural-image datasets ($d=768$) and BiomedCLIP~\cite{zhang2024biomedclip} for medical-image datasets ($d=512$). For CheXpert and ISIC 2020, we apply StratifiedGroupKFold~\cite{scikit-learn} on the combined training and validation sets for hyperparameter tuning and model selection. For all datasets, we used the WtSupCon weights $(\alpha_1,\alpha_2,\alpha_3)=(4,2,1)$. For more details, see~\cref{subsec:albation_hyperparameter_tuning}.
Batch sizes for Waterbirds, ISIC 2020 and CheXpert are 256, 128 and 512, respectively. CheXpert and ISIC 2020 use the LayerNorm-GeLU head with a residual connection ($h, dp$ = (256,0.35) and (512,0.3), respectively) and a \textit{balanced-groups} sampler. For Waterbirds and the toy dataset, we use the BatchNorm-ReLU head (Waterbirds: $h, dp$ = (256,0.3))  and its shallower variant (toy: $h, dp$ = (128,0.3)), with the \textit{ID-paired} sampler.
For all experiments, we train a scikit-learn~\cite{scikit-learn} logistic regression (LR) classifier (lbfgs solver, max\_iter=5000) with a tuned C value, and apply StandardScaler before LR. For Waterbirds, $C=0.001$. For CheXpert and ISIC 2020, $C=0.01$. When fitting on the CheXpert and ISIC 2020 datasets, we used inverse-frequency sample weights based on training groups defined by our predicted spurious labels. Unless otherwise stated, we report results using predicted spurious labels for both the training and validation sets. We use the original prediction procedures for JTT and CA, and apply the prediction procedure proposed in CA to DFR.

 \begin{figure*}
\includegraphics[width=\linewidth]{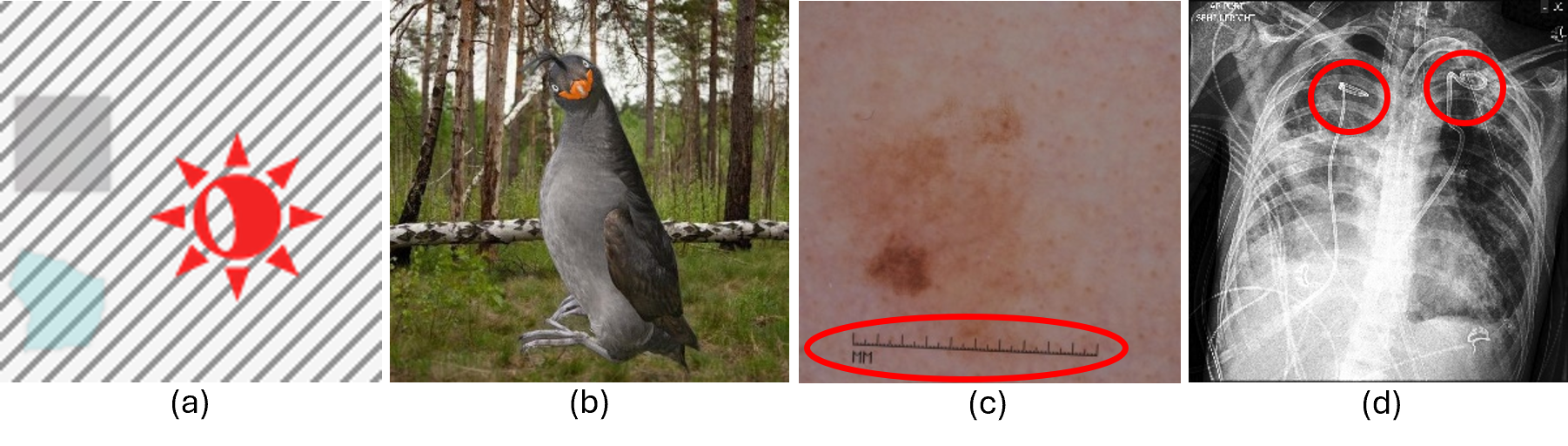}
\caption{\textbf{Examples of spurious attributes across datasets.} (a) Toy example, (b) Waterbirds, (c) ISIC 2020, (d) CheXpert. In (a) and (b), the background serves as the spurious attribute. In (c) and (d), the spurious attribute is circled.} \label{fig:spur_examples}
\end{figure*}













\begin{table*}[!htbp]
\centering
\setlength{\tabcolsep}{4pt} 
\renewcommand{\arraystretch}{1.05} 
\begin{tabular}{lllll}
\toprule


\hiderowcolors
\multicolumn{1}{c}{} &
\multicolumn{4}{c}{\textbf{Waterbirds}} \\
\cline{2-5}
\noalign{\vskip 2pt}
 & \cellcolor{yellow!20} WG & Avg. & Adj. Avg.  & AUC \\
\midrule

Baseline 
    & \cellcolor{yellow!20}58.1
    & 89.0
    & {97.4}
    & 92.5 \\

Baseline-WT 
    & \cellcolor{yellow!20}79.4
    & 88.6 
    & {96.4}
    & 94.0 \\

JTT~\cite{liu2021just}
    & \cellcolor{yellow!20}84.5 $\pm$ 3.0 
    & 89.1 $\pm$ 2.6 
    & 89.7 $\pm$ 4.5 
    & 95.2 $\pm$ 1.2 \\

DFR~\cite{kirichenko2204last}
    & \cellcolor{yellow!20}82.2 $\pm$ 1.1 
    & 87.2 $\pm$ 0.4 
    & 92.4 $\pm$ 0.7 
    & \textbf{97.8} $\pm$ 0.2\\

CA~\cite{zhang2022contrastive}
    & \cellcolor{yellow!20}77.7 $\pm$ 4.8 
    & 86.1 $\pm$ 2.4 
    & 91.9 $\pm$ 3.1 
    & 93.1 $\pm$ 1.8 \\

\midrule

SpurCon \textbf{(Ours)}
    & \cellcolor{yellow!20}\textbf{87.1} $\pm$ 0.8
    & \textbf{94.5} $\pm$ 0.4
    & \textbf{97.5} $\pm$ 0.0
    & 97.4 $\pm$ 0.1\\

\midrule

\multicolumn{1}{c}{} &
\multicolumn{4}{c}{\textbf{CheXpert}} \\
\cline{2-5}
\noalign{\vskip 2pt}
 & \cellcolor{yellow!20} WG & Avg. & Adj. Avg. & AUC \\
\midrule
\rowcolors{1}{lightgray}{white}

Baseline
            & \cellcolor{yellow!20} {$35.6 $}
            & {$63.9 $}
            & {$ 75.9 $}
            & {$ 69.0 $} \\
Baseline-WT
            & \cellcolor{yellow!20} {$71.0 $}
            & {$80.0 $}
            & {$ 84.4 $}
            & {$ 88.8 $} \\

JTT~\cite{liu2021just}
            & \cellcolor{yellow!20} {$57.3\pm 4.6$} & 
            {$71.1\pm 0.8$}& 
            {$ 80.9\pm 1.0$}&
            {$ 78.7\pm0.9 $} \\

DFR~\cite{kirichenko2204last}
            & \cellcolor{yellow!20}  {$58.6\pm1.2 $}
            &{$69.3\pm0.5 $}
            &{$ 76.3\pm0.3 $}
            &{$ 75.5\pm0.7 $} \\

CA~\cite{zhang2022contrastive}
    & \cellcolor{yellow!20} {$ 50.8\pm3.1 $}
    & {$ 67.2 \pm1.5$}
    & {$ 76.2\pm1.4 $}
    & {$72.0 \pm 1.5$} \\
    
\midrule

 SpurCon (\textbf{Ours})  
    & \cellcolor{yellow!20} {$\bm{73.0}\pm0.6 $}
     & {$\bm{80.7}\pm 0.3 $}
     & {$\bm{85.1}\pm 0.5  $}
     & {$ \bm{88.9}\pm0.2 $} \\

\midrule
\hiderowcolors

\multicolumn{1}{c}{} &
\multicolumn{4}{c}{\textbf{ISIC 2020}} \\
\cline{2-5}
\noalign{\vskip 2pt}
 & \cellcolor{yellow!20} WG & Avg. & Adj. Avg. & AUC \\
\midrule
\rowcolors{1}{lightgray}{white}

Baseline            
            & \cellcolor{yellow!20} {$ 0.0$}
            & {$\bm{98.2} $}
            & {$ \bm{98.2} $}
            & {$ 86.3$}    \\

Baseline-WT
            
            & \cellcolor{yellow!20} {$ 58.6$}
            & {$74.8 $}
            & {$ 74.4 $}
            & {$  83.5$}    \\
JTT~\cite{liu2021just}
            & \cellcolor{yellow!20}
            {$ 51.0\pm 2.8$} &
            {${91.4}\pm 0.2$}& 
            {$ {90.9}\pm 0.2$}& 
            {$ \bm{87.2} \pm 0.7$}    \\

DFR~\cite{kirichenko2204last}
            & \cellcolor{yellow!20}  {$ 26.9\pm2.6$}
            &{$61.2\pm2.0 $}
            &{$ 60.1\pm1.9 $}
            &{$85.8\pm0.6$}    \\
 CA~\cite{zhang2022contrastive}
    
     & \cellcolor{yellow!20} {$ 24.6\pm 5.5$}
    & {$ 68.3\pm 6.6$}
    & {$ 66.8\pm6.5 $}
    & {$ 80.5\pm 4.0$}  \\
               
\midrule

 SpurCon (\textbf{Ours})  
     & \cellcolor{yellow!20} {$ \bm{65.0}\pm3.8 $}
     & {$ {80.3}\pm 0.9$}
     & {$ {79.9}\pm 0.9$}
     & {$ {84.8}\pm0.4 $}   \\
\bottomrule

\end{tabular}
\caption{{\textbf {Spurious mitigation results.}} Best results in \textbf{bold}. Across all datasets, \textbf{SpurCon substantially outperforms prior methods in worst-group (WG) accuracy}, demonstrating its effectiveness in mitigating spurious correlations.
 }
 \label{tab:spurcon-results}
\end{table*}

\begin{table}[!htbp]

\centering
\small
\setlength{\tabcolsep}{2.5pt}
\renewcommand{\arraystretch}{0.95}

\begin{tabular}{@{}lccc@{}}
\toprule

& \#Epochs 
& \makecell{$\frac{\text{Time}}{\text{epoch}}$(s)$\downarrow$} 
& Total training time (s)$\downarrow$ \\
\midrule

JTT~\cite{liu2021just}
  
    & {$10(\times2)$} & {$ 137.1\pmt5.3 $} & {$ 2742\pmt107 $}             \\
 DFR~\cite{kirichenko2204last}
  
    & {$75$} & {$ 59.9\pmt 0.8$} & {$4494 \pmt 61$}             \\
 CA~\cite{zhang2022contrastive}
  
    & {$20$} & {$138.3 \pmt 3.5$} & {$2767 \pmt 69$}             \\
\hline

 SpurCon \textbf{(Ours)} 
    & {$35$} & {$\bm{7.9 \pmt 0.6}$} & {$\bm{497 \pmt 36}$}              \\
\bottomrule

\end{tabular}

\caption{\textbf{Training runtime comparison}. SpurCon is \textbf{considerably more computationally efficient} than the evaluated prior methods. Measurements were taken on a single NVIDIA GeForce RTX 3090 GPU and 20 CPU cores. Best result in \textbf{bold}. Obtained on ISIC 2020.
}
\label{runtime-results}
\end{table}
\begin{figure}
\includegraphics[width=\linewidth]{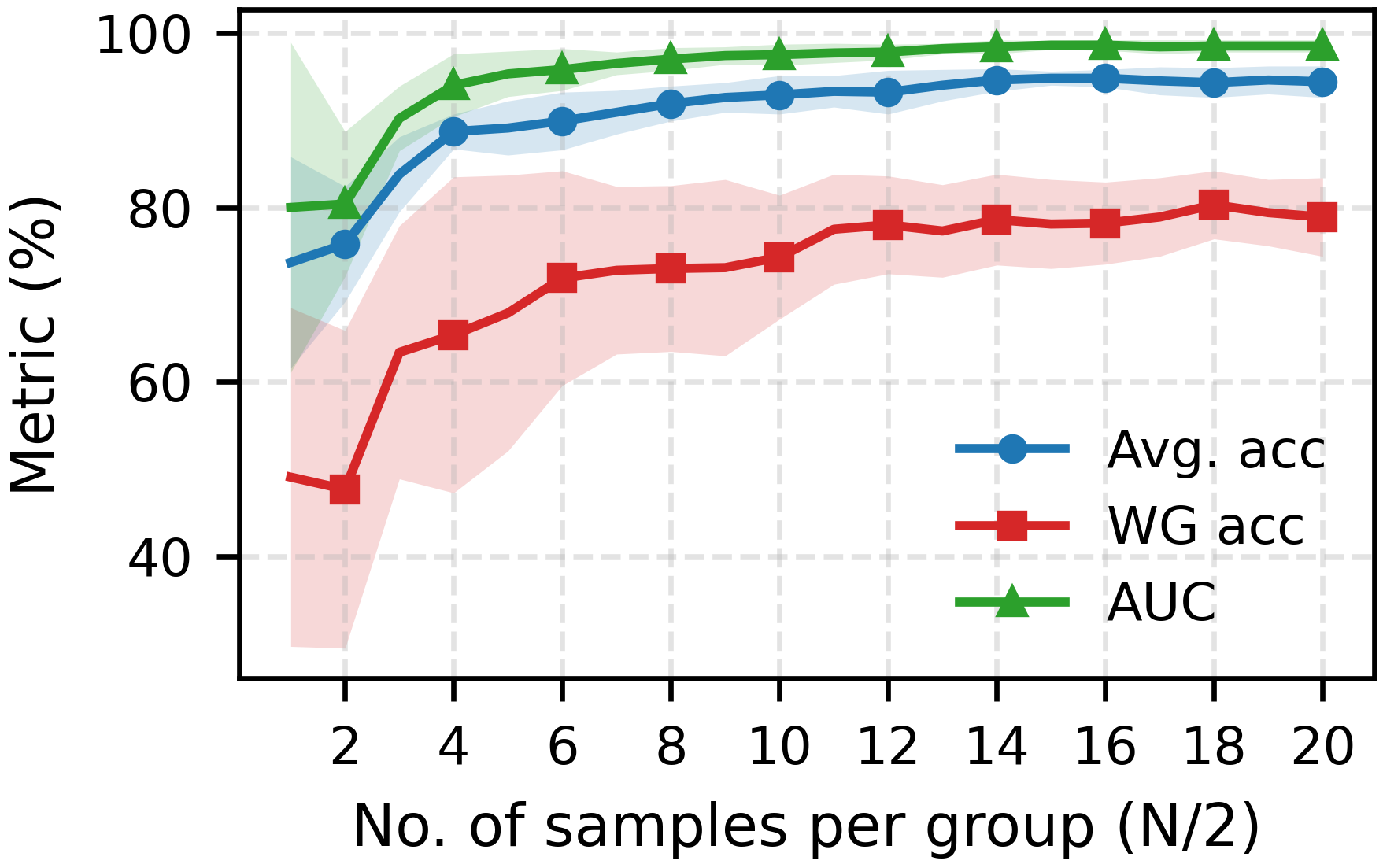}
\caption{\textbf{Performance vs. $\#$ annotated samples.} Metrics largely plateau with few samples. We select $\bf{N/2=10}$ to balance annotation efficiency and performance. Obtained on Waterbirds.} \label{graph:tuning_expert_n}
\end{figure}

\begin{table*}[t]

\centering
\begin{tabular}{l|c|c|cc|cc}
\hline
    & \makecell{baseline-WT}
    & \makecell{Original}
    & \multicolumn{2}{c|}{Expert annotation mistakes}
    & \multicolumn{2}{c}{Estimated annotations mistakes} \\
    \cline{4-7}
 & 
 & 
 & \makecell{$10\%$}
 & \makecell{$20\%$}
 & \makecell{$10\%$}
 & \makecell{$20\%$} \\
 
 \hline
\rowcolor{yellow!20} WG  &   {$58.6$}
     &{$ 65.0\pm3.8$}
     &{$ 64.0\pm4.7$}
     & {$ 58.9\pm4.6 $}
     &{$ 60.8\pm 3.8$}
     & {$ 54.7\pm4.2 $}\\
 \rowcolor{lightgray} Avg. &  {$ 74.8$}
      &{$ 80.3\pm0.9$}
      &{$ 80.3\pm1.3 $} 
      & {$ 79.2\pm 1.9$}
      &{$ 81.2\pm1.9 $} 
      & {$ 80.0\pm 1.3$}  \\
\rowcolor{white} Adj. Avg. & {$74.4$}
           &{$79.9\pm0.9 $}
           &{$ 79.8\pm 1.3$} 
           & {$ {78.7\pm1.9} $}
           &{$ 80.7\pm 2.0$} 
           & {$ {79.2\pm1.3} $} \\

\rowcolor{lightgray} AUC & {$83.5$}
     & {$84.8\pm0.4 $}
     &{$ 84.8\pm 1.0 $} 
     & {$ 84.9\pm 1.1$}
     &{$ 84.8\pm 1.0 $} 
     & {$ 85.8\pm 0.5$} \\
\hline
\end{tabular}
\caption{\textbf{Robustness analysis - Pick-and-Predict.} Under moderate corruption, Pick-and-Predict remains above the baseline. Higher WG accuracy under corrupt expert annotations than corrupt estimated labels indicates the robustness of the estimation process. In each case, $p\%$ of labels are corrupted within each $(\text{class},\text{spurious})$. Obtained on ISIC 2020.}
\label{pred_spur_robustness}
\end{table*}

\begin{table*}
\centering
\begin{tabular}{l|c|c|c|c|c}
\hline
 & SupCon & \multicolumn{4}{c}{WtSupCon variants \textbf{(Ours)}} \\
\cline{3-6}
 & & \makecell{Ablation 1\\$\alpha_1=\alpha_2=\alpha_3$} 
 & \makecell{Ablation 2\\$\alpha_1=\alpha_2>\alpha_3$} 
 & \makecell{Ablation 3\\$\alpha_1>\alpha_2=\alpha_3$} 
 & \makecell{Ablation 4\\$\alpha_1>\alpha_2>\alpha_3$} \\
\hline
\rowcolor{yellow!20} WG  
&{$ 61.2\pmt1.5$}
&{$ 60.3\pmt 2.3$} 
& {$ 58.1\pmt2.4 $} 
&{$78.5 \pmt3.4 $}
&{$\bm{80.7\pmt2.0} $}\\

\rowcolor{white} Avg. 
&  {$ 84.3\pmt0.5$}
&{$ 83.9\pmt0.8 $} 
& {$ 82.9\pmt 1.2$} &{$ 91.6\pmt0.9 $}&{$\bm{92.0 \pmt0.5 }$} \\
\rowcolor{lightgray} Adj. Avg. &  {$98.0\pmt0.1 $}&{$ 97.9\pmt 0.2$} & {$ \bm{98.1\pmt0.1} $} &{$97.0\pmt 0.4$}&{$ 96.6\pmt0.3 $} \\

\rowcolor{white} AUC &  {$93.6\pmt0.5 $}&{$ 93.3\pmt 0.6 $} & {$ 92.7\pmt 0.8$} &{$97.5 \pmt0.5 $}&{$\bm{97.8\pmt0.1} $}\\
\hline
\end{tabular}
\caption{\textbf{WtSupCon weights relations.} Evaluation against standard SupCon and multiple WtSupCon weight settings on the toy dataset. Ablation~4 achieves the highest worst-group (WG) accuracy, validating our choice of this weighting scheme for SpurCon. Results obtained using true spurious labels. Best results in \bf bold.}
\label{tab:toy-results}
\end{table*}

\begin{table}[t]
\centering
\setlength{\tabcolsep}{1.5pt}

\begin{subtable}{\linewidth}
\centering
\begin{tabular}{@{}cccccccccc@{}}
\toprule
& & & & \multicolumn{3}{c}{WG} & \multicolumn{3}{c}{Avg} \\
\cmidrule(lr){5-7} \cmidrule(lr){8-10}
Rank & $\alpha_1$ & $\alpha_2$ & $\alpha_3$
& $\mu$ & $\sigma$ & $\mu-\sigma$
& $\mu$ & $\sigma$ & $\mu-\sigma$ \\
\midrule
1 & 4     & 2 & 1 & 67.5 & 2.7 & \bf{64.8} & 84.2 & 1.6 & \bf{82.6} \\

2 & 50000 & 2 & 1 & \bf{67.7} & 3.1 & 64.6 & 84.3 & 2.0 & 82.4 \\

3 & 1000  & 2 & 1 & 67.1 & 3.3 & 63.9 & \bf{85.4} & 2.9 & 82.5 \\
\bottomrule
\end{tabular}

\caption{}
\label{tab:isic_top3_and_search}
\end{subtable}

\vspace{0.6em}
\begin{subtable}{\linewidth}
\centering
\begin{tabular}{@{}cl@{}}
\toprule
Parameter & Evaluated values \\
\midrule
$\alpha_1$ & $\{4,20,100,500,$ \\
           & $\phantom{\{}1000,5000,10000,$ \\
           & $\phantom{\{}50000,100000\}$ \\
$\alpha_2$ & $\{2,10,100,1000\}$ \\
$\alpha_3$ & $\{1\}$ \\
\midrule
Total & 15 combinations \\
\bottomrule
\end{tabular}

\caption{}
\label{tab:combinations_weights}
\end{subtable}
\caption{(a): Top configurations results. $\bf{(\alpha_1, \alpha_2, \alpha_3)=(4,2,1)}$ \textbf{achieves the highest rank} according to $\mu-\sigma$ values. Obtained on ISIC 2020. (b): Evaluated hyperparameter values.}
\label{tab:hyperparameter_tuning}
\end{table}

\subsection{Results}
 All results are reported with standard deviation over five random seeds.

\noindent\textbf{SpurCon analysis.} We compare SpurCon on the three datasets against several methods: Baseline, a logistic-regression classifier trained directly on the raw image embeddings, Baseline-WT, the same logistic-regression classifier, including inverse-frequency sample weights during fitting, Just Train Twice (JTT)~\cite{liu2021just}, Deep Feature Reweighting (DFR)~\cite{kirichenko2204last}, and Contrastive Adapters (CA)~\cite{zhang2022contrastive}, which is the prior approach most closely related to ours. The pretrained image encoder used in all CA experiments is CLIP~\cite{radford2021learning}, and for DFR and JTT, we train an ImageNet-pretrained ResNet-50 model on each examined dataset, following their original implementation on Waterbirds. Note that the baseline methods are reported without standard deviations because the deterministic solver used produces identical results across runs when trained on the same features and labels.
As can be seen in~\cref{tab:spurcon-results}, SpurCon significantly outperforms prior methods on Waterbirds across all metrics except AUC, where it trails DFR by a small margin. SpurCon performs best on all reported metrics for CheXpert, where the competition is mostly with the baseline-WT results. Baseline-WT assigns inverse-frequency weights based on training groups formed using the spurious labels predicted by our few-shot method. This helps the simple classifier handle the complex group distribution of the CheXpert, leading to the second-best performance. For ISIC 2020, although the baseline achieves higher average accuracies than SpurCon, its worst-group accuracy is zero. This indicates a strong bias toward the majority groups, which are substantially larger than the minority groups in this dataset (see Supp.). Among the examined methods, SpurCon achieves the best balance between average and worst-group accuracy. These results highlight the effectiveness of SpurCon in the medical domain.

\noindent\textbf{Time comparison.} \Cref{runtime-results} shows that, in the predicted spurious-label setting, SpurCon is order-of-magnitude faster than the other methods in both time per epoch and total training time. SpurCon is $\sim$17$\times$ faster per epoch, $\sim$5.5$\times$ faster in total training time than JTT and CA, $\sim$7$\times$ faster per epoch and $\sim$9$\times$ faster in total training time than DFR. Note that JTT time per epoch is amortized over both training stages, the initial ERM run and the final upweighted JTT run. The results demonstrate that SpurCon combines improved robustness with substantially lower computational cost.

\noindent\textbf{Spurious labels prediction analysis.} We apply our few-shot spurious-label estimation method to Waterbirds, CheXpert and ISIC 2020. To construct spurious-label prototypes with all groups $\mathcal{G}$ balanced, we manually annotate 10 samples per group, equivalently 20 samples per spurious label, for each dataset. The number of annotated samples per spurious label, $(N)$, is a tunable hyperparameter, as shown in~\cref{graph:tuning_expert_n}. We choose $(N/2=10)$, since performance changes only marginally beyond this point. Keeping $N$ small also preserves scalability and annotation efficiency, particularly for datasets with multiple target classes and spurious attributes.
We then use these annotations to estimate spurious labels for the remaining training and validation samples. Our approach achieves AUCs of $95.4\%$ (Waterbirds), $97.3\%$ (CheXpert) and $80.9\%$ (ISIC 2020).

\noindent\textbf{Robustness to noisy spurious labels.}
Spurious cues may sometimes be ambiguous. Therefore, the robustness of SpurCon to noisy spurious labels was examined. Specifically, either the expert annotations before spurious-label estimation or the estimated spurious labels after estimation were progressively corrupted, as presented in \cref{pred_spur_robustness}. On both noise levels, corrupting in expert-level achieves higher WG accuracy than in estimation-lebel, while the other metrics remain approximately similar to those of the original experiment without label noise. This suggests that the few-shot estimator is robust to moderate annotation noise. Moreover, even when $20\%$ expert-label mistakes, our method outperforms the baseline, indicating that the overall method remains robust to noisy spurious labels.

\begin{figure}[!ht]
\centering

\begin{tikzpicture}
\begin{axis}[
    width=\columnwidth,
    height=2.6cm,
    xmin=61,
    xmax=69,
    title={WG-based metrics},
    symbolic y coords={{$\mu$}, {$\mu-\sigma$}},
    ytick=data,
    y dir=reverse,
    tick label style={font=\small},
    label style={font=\small},
    title style={font=\small},
    axis x line*=bottom,
    axis y line*=left,
    xmajorgrids,
    grid style={dashed},
    enlarge y limits=0.25
]

\addplot[thick, forget plot] coordinates { (64.8,{$\mu$}) (67.7,{$\mu$}) };
\addplot[thick, forget plot] coordinates { (62.1,{$\mu-\sigma$}) (65.0,{$\mu-\sigma$}) };

\addplot[
    only marks,
    mark=o,
    mark size=2.2pt
] coordinates {
    (64.8,{$\mu$})
    (62.1,{$\mu-\sigma$})
};

\addplot[
    only marks,
    mark=*,
    mark size=2.2pt
] coordinates {
    (67.7,{$\mu$})
    (65.0,{$\mu-\sigma$})
};

\end{axis}
\end{tikzpicture}

\vspace{0.4em}

\begin{tikzpicture}
\begin{axis}[
    width=\columnwidth,
    height=2.6cm,
    xmin=80,
    xmax=88,
    xlabel={Validation performance (\%)},
    title={Average-based metrics},
    symbolic y coords={{$\mu$}, {$\mu-\sigma$}},
    ytick=data,
    y dir=reverse,
    tick label style={font=\small},
    label style={font=\small},
    title style={font=\small},
    axis x line*=bottom,
    axis y line*=left,
    xmajorgrids,
    grid style={dashed},
    enlarge y limits=0.25
]

\addplot[thick, forget plot] coordinates { (83.4,{$\mu$}) (87.1,{$\mu$}) };
\addplot[thick, forget plot] coordinates { (81.0,{$\mu-\sigma$}) (84.1,{$\mu-\sigma$}) };

\addplot[
    only marks,
    mark=o,
    mark size=2.2pt
] coordinates {
    (83.4,{$\mu$})
    (81.0,{$\mu-\sigma$})
};

\addplot[
    only marks,
    mark=*,
    mark size=2.2pt
] coordinates {
    (87.1,{$\mu$})
    (84.1,{$\mu-\sigma$})
};

\end{axis}
\end{tikzpicture}

\caption{\textbf{Variation across combinations.} Performance across the 15 weight combinations. Each segment spans the worst-to-best result. \textbf{Low variance across weight combinations indicates robustness to weight selection.} Obtained on ISIC 2020.}
\label{fig:isic_spread}
\end{figure}
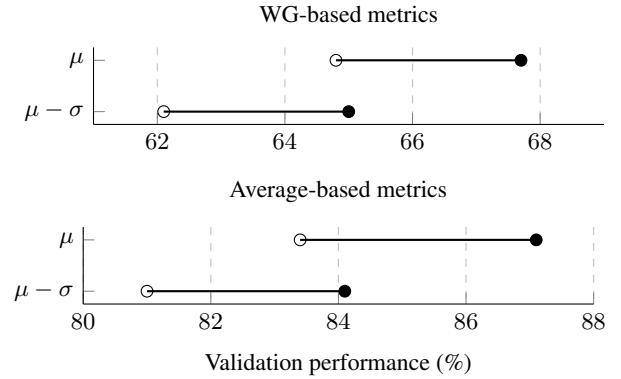

\subsection{Ablation study}
\label{subsec:albation_hyperparameter_tuning} 
\noindent\textbf{Weights relations.}
We assess the WtSupCon loss on our toy dataset by comparing it to the standard SupCon loss and by ablating the three positive-pair sets (1--3 in~\cref{weight-sets}), progressively adding them back until the full configuration is restored. \Cref{tab:toy-results} shows that $\alpha_1$ is the main driver of worst-group accuracy, as Ablation~3 outperforms Ablations~1 and~2 by a large margin. Ablation~4 achieves the highest worst-group accuracy, supporting our claim that this set-weighting ratio is key to improving robustness to spurious correlations.

\noindent\textbf{Hyperparameter tuning.}
 The WtSupCon weights $(\alpha_1, \alpha_2, \alpha_3)$ are selected in two stages. First, we use a coarse initialization in which the weights are set inversely proportional to corresponding group population in the training set. We then tune the weights over a fixed set of 15 predefined configurations, chosen to 
represent different relative weighting regimes between the three set weights (see~\cref{tab:combinations_weights}). The final configuration is selected on the ISIC 2020 validation set using a rank-based stability criterion. Specifically, we first restrict the candidates to the top three configurations according to mean WG accuracy. Then, we select the configuration with the highest WG mean-minus-standard-deviation score; 
average-accuracy mean-minus-standard-deviation is used as an additional stability 
check. This criterion favors configurations that achieve high WG accuracy while 
remaining stable across seeds and preserving competitive average accuracy. Accordingly, we select $(\alpha_1,\alpha_2,\alpha_3)=(4,2,1)$. Results are reported in~\cref{tab:isic_top3_and_search}.
Performance varied only mildly across the evaluated configurations, suggesting that the method is not sensitive to the fine-grained 
choice of weights (see~\cref{fig:isic_spread}). 
\section{Conclusion}
\label{sec:conclusions}
This paper introduces SpurCon, a lightweight method that presents a new supervised contrastive objective, WtSupCon, that uses available metadata (\eg, patient ID) together with predicted spurious labels to discourage strong pretrained embeddings from relying on spurious attributes. Reliable predicted spurious labels are estimated using our proposed simple few-shot approach, which combines a powerful pretrained image encoder and minimal expert guidance.
Experimental evaluations on three publicly available datasets, including two medical imaging datasets, demonstrate that SpurCon outperforms existing spurious-mitigation methods, delivering better performance metrics while also reducing training time.
{
    \small
    \bibliographystyle{ieeenat_fullname}
    \bibliography{main}

@String(AAAI = {AAAI})

@article{zhang2022correct,
  title={Correct-n-contrast: A contrastive approach for improving robustness to spurious correlations},
  author={Zhang, Michael and Sohoni, Nimit S and Zhang, Hongyang R and Finn, Chelsea and R{\'e}, Christopher},
  journal={arXiv preprint arXiv:2203.01517},
  year={2022}
}

@article{zhang2022contrastive,
  title={Contrastive adapters for foundation model group robustness},
  author={Zhang, Michael and R{\'e}, Christopher},
  journal={Advances in Neural Information Processing Systems},
  volume={35},
  pages={21682--21697},
  year={2022}
}

@inproceedings{vu2021medaug,
  title={Medaug: Contrastive learning leveraging patient metadata improves representations for chest x-ray interpretation},
  author={Vu, Yen Nhi Truong and Wang, Richard and Balachandar, Niranjan and Liu, Can and Ng, Andrew Y and Rajpurkar, Pranav},
  booktitle={Machine Learning for Healthcare Conference},
  pages={755--769},
  year={2021},
  organization={PMLR}
}

@article{khosla2020supervised,
  title={Supervised contrastive learning},
  author={Khosla, Prannay and Teterwak, Piotr and Wang, Chen and Sarna, Aaron and Tian, Yonglong and Isola, Phillip and Maschinot, Aaron and Liu, Ce and Krishnan, Dilip},
  journal={Advances in neural information processing systems},
  volume={33},
  pages={18661--18673},
  year={2020}
}

@article{sagawa2019distributionally,
  title={Distributionally robust neural networks for group shifts: On the importance of regularization for worst-case generalization},
  author={Sagawa, Shiori and Koh, Pang Wei and Hashimoto, Tatsunori B and Liang, Percy},
  journal={arXiv preprint arXiv:1911.08731},
  year={2019}
}

@article{rotemberg2021patient,
  title={A patient-centric dataset of images and metadata for identifying melanomas using clinical context},
  author={Rotemberg, Veronica and Kurtansky, Nicholas and Betz-Stablein, Brigid and Caffery, Liam and Chousakos, Emmanouil and Codella, Noel and Combalia, Marc and Dusza, Stephen and Guitera, Pascale and Gutman, David and others},
  journal={Scientific data},
  volume={8},
  number={1},
  pages={34},
  year={2021},
  publisher={Nature Publishing Group UK London}
}

@inproceedings{radford2021learning,
  title={Learning transferable visual models from natural language supervision},
  author={Radford, Alec and Kim, Jong Wook and Hallacy, Chris and Ramesh, Aditya and Goh, Gabriel and Agarwal, Sandhini and Sastry, Girish and Askell, Amanda and Mishkin, Pamela and Clark, Jack and others},
  booktitle={International conference on machine learning},
  pages={8748--8763},
  year={2021},
  organization={PmLR}
}

@article{bevan2021skin,
  title={Skin deep unlearning: Artefact and instrument debiasing in the context of melanoma classification},
  author={Bevan, Peter J and Atapour-Abarghouei, Amir},
  journal={arXiv preprint arXiv:2109.09818},
  year={2021}
}

@article{nam2022spread,
  title={Spread spurious attribute: Improving worst-group accuracy with spurious attribute estimation},
  author={Nam, Junhyun and Kim, Jaehyung and Lee, Jaeho and Shin, Jinwoo},
  journal={arXiv preprint arXiv:2204.02070},
  year={2022}
}

@article{scikit-learn,
  title={Scikit-learn: Machine Learning in {P}ython},
  author={Pedregosa, Fabian and Varoquaux, Ga{\"e}l and Gramfort, Alexandre and Michel, Vincent and Thirion, Bertrand and Grisel, Olivier and Blondel, Mathieu and Prettenhofer, Peter and Weiss, Ron and Dubourg, Vincent and Vanderplas, Jake and Passos, Alexandre and Cournapeau, David and Brucher, Matthieu and Perrot, Matthieu and Duchesnay, {\'E}douard},
  journal={Journal of Machine Learning Research},
  volume={12},
  pages={2825--2830},
  year={2011}
}

@article{lekadir2025future,
  title={FUTURE-AI: international consensus guideline for trustworthy and deployable artificial intelligence in healthcare},
  author={Lekadir, Karim and Frangi, Alejandro F and Porras, Antonio R and Glocker, Ben and Cintas, Celia and Langlotz, Curtis P and Weicken, Eva and Asselbergs, Folkert W and Prior, Fred and Collins, Gary S and others},
  journal={bmj},
  volume={388},
  year={2025},
  publisher={British Medical Journal Publishing Group}
}

@inproceedings{saab2022reducing,
  title={Reducing reliance on spurious features in medical image classification with spatial specificity},
  author={Saab, Khaled and Hooper, Sarah and Chen, Mayee and Zhang, Michael and Rubin, Daniel and R{\'e}, Christopher},
  booktitle={Machine Learning for Healthcare Conference},
  pages={760--784},
  year={2022},
  organization={PMLR}
}

@article{ye2024spurious,
  title={Spurious correlations in machine learning: A survey},
  author={Ye, Wenqian and Zheng, Guangtao and Cao, Xu and Ma, Yunsheng and Zhang, Aidong},
  journal={arXiv preprint arXiv:2402.12715},
  year={2024}
}

@article{sohoni2021barack,
  title={Barack: Partially supervised group robustness with guarantees},
  author={Sohoni, Nimit S and Sanjabi, Maziar and Ballas, Nicolas and Grover, Aditya and Nie, Shaoliang and Firooz, Hamed and R{\'e}, Christopher},
  journal={arXiv preprint arXiv:2201.00072},
  year={2021}
}

@article{du2023shortcut,
  title={Shortcut learning of large language models in natural language understanding},
  author={Du, Mengnan and He, Fengxiang and Zou, Na and Tao, Dacheng and Hu, Xia},
  journal={Communications of the ACM},
  volume={67},
  number={1},
  pages={110--120},
  year={2023},
  publisher={ACM New York, NY, USA}
}

@inproceedings{irvin2019chexpert,
  title={Chexpert: A large chest radiograph dataset with uncertainty labels and expert comparison},
  author={Irvin, Jeremy and Rajpurkar, Pranav and Ko, Michael and Yu, Yifan and Ciurea-Ilcus, Silviana and Chute, Chris and Marklund, Henrik and Haghgoo, Behzad and Ball, Robyn and Shpanskaya, Katie and others},
  booktitle={Proceedings of the AAAI conference on artificial intelligence},
  volume={33},
  number={01},
  pages={590--597},
  year={2019}
}

@article{obermeyer2019dissecting,
  title={Dissecting racial bias in an algorithm used to manage the health of populations},
  author={Obermeyer, Ziad and Powers, Brian and Vogeli, Christine and Mullainathan, Sendhil},
  journal={Science},
  volume={366},
  number={6464},
  pages={447--453},
  year={2019},
  publisher={American Association for the Advancement of Science}
}

@article{waller2022applications,
  title={Applications and challenges of artificial intelligence in diagnostic and interventional radiology},
  author={Waller, Joseph and O’connor, Aisling and Raafat, Eleeza and Amireh, Ahmad and Dempsey, John and Martin, Clarissa and Umair, Muhammad},
  journal={Polish journal of radiology},
  volume={87},
  number={1},
  pages={113--117},
  year={2022},
  publisher={Termedia}
}

@article{huang2023self,
  title={Self-supervised learning for medical image classification: a systematic review and implementation guidelines},
  author={Huang, Shih-Cheng and Pareek, Anuj and Jensen, Malte and Lungren, Matthew P and Yeung, Serena and Chaudhari, Akshay S},
  journal={NPJ Digital Medicine},
  volume={6},
  number={1},
  pages={74},
  year={2023},
  publisher={Nature Publishing Group UK London}
}

@article{kirichenko2204last,
  title={Last layer re-training is sufficient for robustness to spurious correlations, 2023},
  author={Kirichenko, Polina and Izmailov, Pavel and Wilson, Andrew Gordon},
  journal={URL https://arxiv. org/abs/2204.02937},
  volume={9},
  pages={26}
}

@inproceedings{liu2021just,
  title={Just train twice: Improving group robustness without training group information},
  author={Liu, Evan Z and Haghgoo, Behzad and Chen, Annie S and Raghunathan, Aditi and Koh, Pang Wei and Sagawa, Shiori and Liang, Percy and Finn, Chelsea},
  booktitle={International Conference on Machine Learning},
  pages={6781--6792},
  year={2021},
  organization={PMLR}
}

@article{zhang2024biomedclip,
  title={A Multimodal Biomedical Foundation Model Trained from Fifteen Million Image–Text Pairs},
  author={Sheng Zhang and Yanbo Xu and Naoto Usuyama and Hanwen Xu and Jaspreet Bagga and Robert Tinn and Sam Preston and Rajesh Rao and Mu Wei and Naveen Valluri and Cliff Wong and Andrea Tupini and Yu Wang and Matt Mazzola and Swadheen Shukla and Lars Liden and Jianfeng Gao and Angela Crabtree and Brian Piening and Carlo Bifulco and Matthew P. Lungren and Tristan Naumann and Sheng Wang and Hoifung Poon},
  journal={NEJM AI},
  year={2024},
  volume={2},
  number={1},
  doi={10.1056/AIoa2400640},
  url={https://ai.nejm.org/doi/full/10.1056/AIoa2400640}
}

@article{vollmer2020machine,
  title={Machine learning and artificial intelligence research for patient benefit: 20 critical questions on transparency, replicability, ethics, and effectiveness},
  author={Vollmer, Sebastian and Mateen, Bilal A and Bohner, Gergo and Kir{\'a}ly, Franz J and Ghani, Rayid and Jonsson, Pall and Cumbers, Sarah and Jonas, Adrian and McAllister, Katherine SL and Myles, Puja and others},
  journal={bmj},
  volume={368},
  year={2020},
  publisher={British Medical Journal Publishing Group}
}
}

\makeatletter

\let\maketitle\savedmaketitle
\let\@maketitle\savedinternalmaketitle

\gdef\@title{%
SpurCon: Weighted Supervised Contrastive Learning for Mitigating Spurious Cues in Medical Imaging\\
{\large --- Supplementary Material ---}%
}
\gdef\@author{%
Shenhav Nadir\textsuperscript{*} \quad
Meir Yossef Levi \quad
Eyal Gofer\textsuperscript{\dag} \quad
Guy Gilboa\textsuperscript{\dag}\\[1ex]
Viterbi Faculty of Electrical and Computer Engineering, Technion - Israel Institute of Technology\\
{\small\textsuperscript{*}\texttt{shenhav.n@campus.technion.ac.il}
}}
\makeatother

\maketitle
\AddToHookNext{shipout/foreground}{%
  \put(54,-758){%
    \makebox[0pt][l]{%
      \scriptsize\textsuperscript{\dag} Joint supervision.
    }%
  }%
}
\appendix

\section{Datasets}

\subsection{Toy dataset}
\begin{figure*}[!htbp]
\centering
\includegraphics[width=\textwidth]{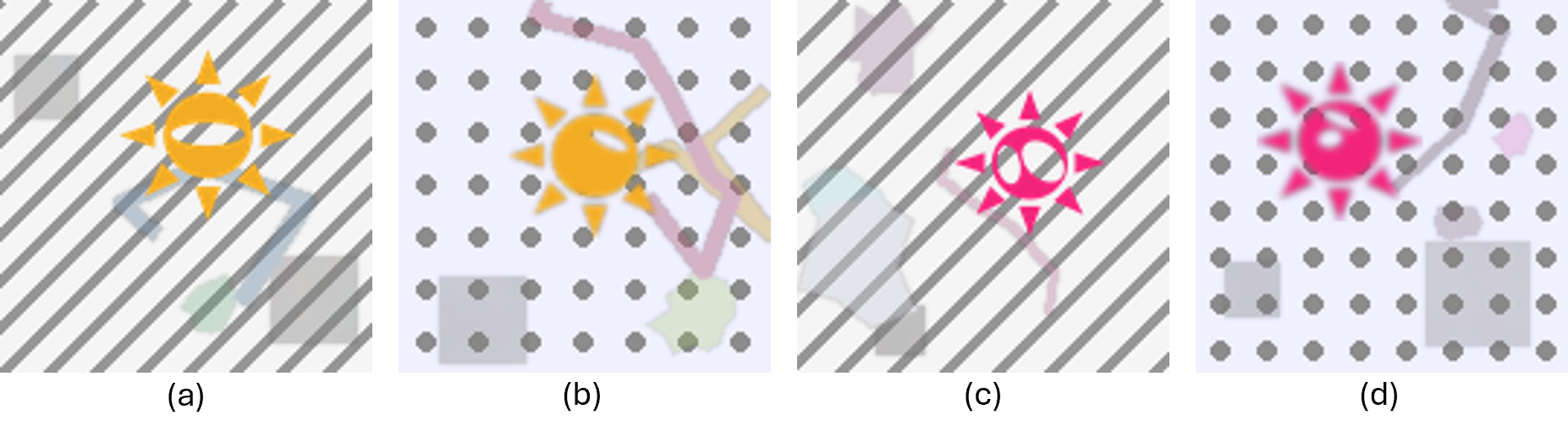}
\caption{\textbf{Toy dataset example.} Images (a) and (b) belong to the class $y=one \:hole=0$ and share the same ID, orange, while differing in their spurious background attribute. Images (c) and (d) belong to the class $y=two \:holes=1$ and share the same ID, pink, while also differing in the background. The examples illustrate variations in template, including hole size and location, as well as scale, jitter, and random background objects.}
\label{fig:toy_example}
\end{figure*}

A binary classification dataset created by us, consisting of images with classes $\mathcal{Y}=\{\text{one hole},\text{two holes}\},$ encoded as $\{0,1\}$, respectively.
There are two background types used as a spurious attribute: $S=\{\text{stripes},\text{dots}\}$, encoded as $\{0,1\}$, respectively.
The dataset uses five templates per class, with variations in scale, jitter, blurring and random small background objects, and $100$ different colors, which serve as IDs.
The train/validation/test splits comprise $7,000/1,000/2,000$ samples; the training set includes (class, spurious) combinations of $(0,0)=47.5\%$, $(0,1)=2.5\%$, $(1,0)=2.5\%$, and $(1,1)=47.5\%$, the validation set uses intermediate proportions, with combinations of $(0,0)=35\%$, $(0,1)=15\%$, $(1,0)=15\%$, and $(1,1)=35\%$, and the test set is balanced between the four groups. \Cref{fig:toy_example} provides examples from the toy dataset.

\subsection{\texorpdfstring{ISIC 2020~\cite{rotemberg2021patient}}{ISIC 2020}
}
A binary classification dataset with classes $\mathcal{Y}=\{\text{benign}, \text{malignant} \}$, encoded as $\{0,1\}$, respectively. We combine metadata from~\cite{bevan2021skin,rotemberg2021patient} to obtain patient IDs and a binary spurious attribute indicating ruler presence, $S=\{\text{without\_ruler},\text{with\_ruler}\}$, encoded as $\{0,1\}$, respectively. Following~\cite{bevan2021skin}, the images are center-cropped and resized to $256\times256$, yielding 32,692 images.
The dataset distribution based on spurious labels predicted by our few-shot estimation method is reported in~\Cref{tab:isic_pred_dataset_distribution}. Note the substantial class imbalance, with the number of benign samples greatly exceeding that of malignant samples. To compute Adj. Avg., we use the true group proportions in the combined training and validation sets: $(0,0)=22617$, $(0,1)=3264$, $(1,0)=352$, $(1,1)=114$.

\begin{table}[!htbp]
\centering

\setlength{\tabcolsep}{5pt}
\renewcommand{\arraystretch}{1.15}
\begin{tabular}{lrrrrr}
\toprule
\textbf{Split}
& \(\boldsymbol{(0,0)}\)
& \(\boldsymbol{(0,1)}\)
& \(\boldsymbol{(1,0)}\)
& \(\boldsymbol{(1,1)}\)
& \textbf{Total} \\
\midrule
Train      & 17527 & 3427 & 254 & 135 & 21343 \\
Validation & 4156 & 771 & 53 & 24 & 5004 \\
Test       & 5605 & 625 & 88 & 27 & 6345 \\
\midrule
Total      & 27288 & 4823 & 395 & 186 & 32692 \\
\bottomrule
\end{tabular}
\caption{\textbf{ISIC 2020 distribution.} Each group is defined as $(y,s)$, where $y$ is the class label and $s$ is the spurious label. Predicted spurious labels are used for the training and validation sets, true spurious labels are used for the test set. \textbf{Note the severe class imbalance, with substantially more benign than malignant samples.}}
\label{tab:isic_pred_dataset_distribution}
\end{table}

\subsection{\texorpdfstring{CheXpert - Pneumothorax~\cite{irvin2019chexpert}}{Chexpert}
}
A large multi-label chest X-ray classification dataset. We construct a binary classification task using the \textit{No Finding} and \textit{Pneumothorax} labels, defining $\mathcal{Y}=\{\text{No Finding}, \text{Pneumothorax} \}$, encoded as $\{0,1\}$, respectively. We use the \textit{Support Devices} column as the binary spurious attribute, defining $\mathcal{S}=\{\text{without\_support\_devices},\text{with\_support\_devices}\}$, encoded as $\{0,1\}$, respectively. The metadata also includes patient IDs. Within each class, the training and validation sets are highly imbalanced between the two groups defined by the predicted spurious labels ($\approx98\%/2\%$), whereas the test set is balanced within each class. In total, 36,679 X-ray scans were used. The dataset distribution based on spurious labels predicted by our few-shot estimation method is reported in~\Cref{tab:chexpert_pred_dataset_distribution}. To compute Adj. Avg., we use the true group proportions in the combined training and validation sets: $(0,0)=17271$, $(0,1)=221$, $(1,0)=690$, $(1,1)=16018$.

\begin{table}[!htbp]
\centering

\setlength{\tabcolsep}{5pt}
\renewcommand{\arraystretch}{1.15}
\begin{tabular}{lrrrrr}
\toprule
\textbf{Split}
& \(\boldsymbol{(0,0)}\)
& \(\boldsymbol{(0,1)}\)
& \(\boldsymbol{(1,0)}\)
& \(\boldsymbol{(1,1)}\)
& \textbf{Total} \\
\midrule
Train      & 13733 & 277 & 246 & 13147 & 27403 \\
Validation & 3415 & 67 & 61 & 3254 & 6797 \\
Test       & 416 & 416 & 803 & 844 & 2479 \\
\midrule
Total      & 17564 & 760 & 1110 & 17245 & 36679 \\
\bottomrule
\end{tabular}
\caption{\textbf{CheXpert distribution.} Each group is defined as $(y,s)$, where $y$ is the class label and $s$ is the spurious label. Predicted spurious labels are used for the training and validation sets, true spurious labels are used for the test set.}
\label{tab:chexpert_pred_dataset_distribution}
\end{table}

\section{Motivation for metadata usage}
\begin{figure*}[!htbp]
\includegraphics[width=\textwidth]{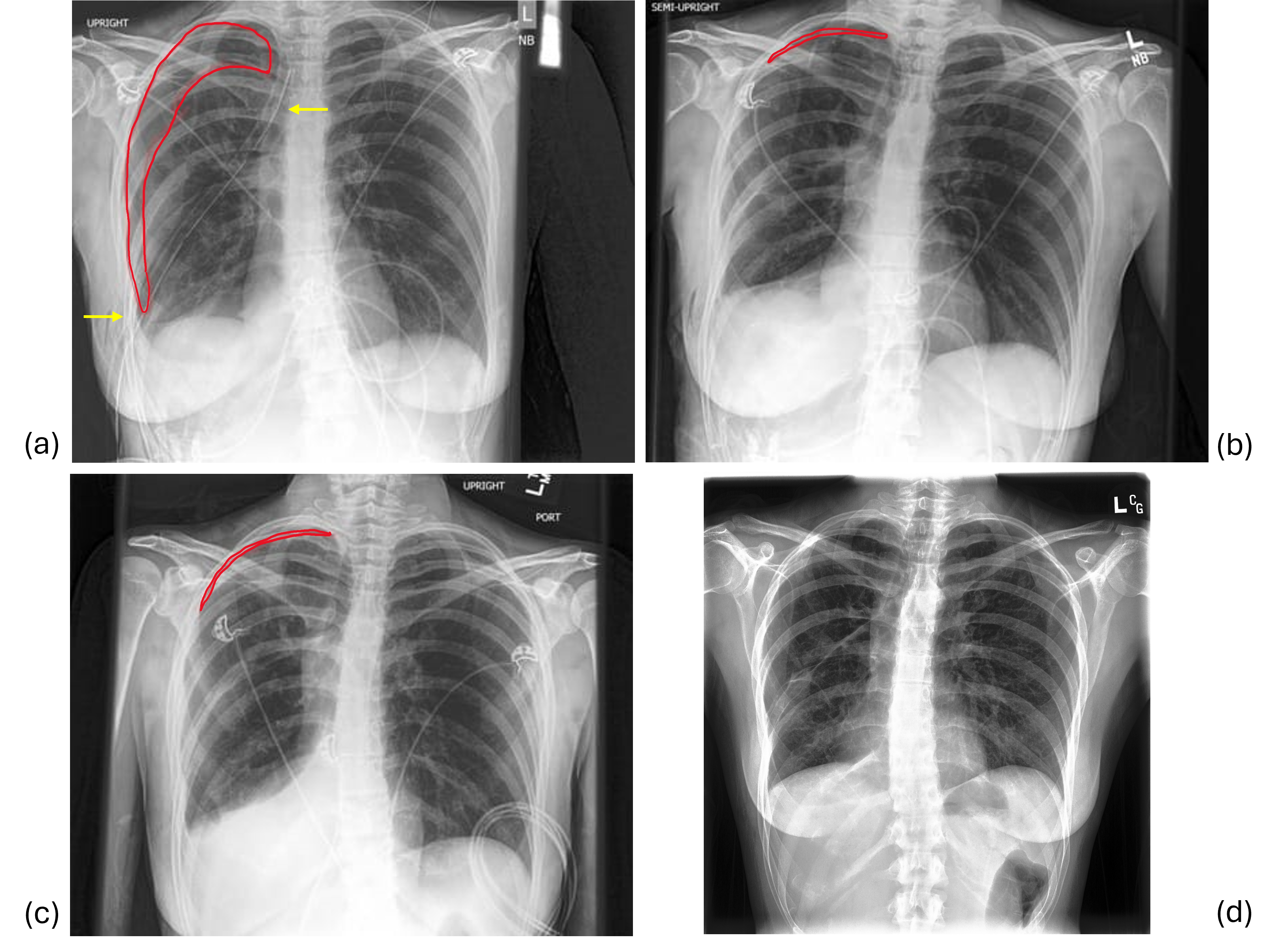}
\caption{\textbf{Comparison of images from the same patient.} All four chest X-ray images were acquired from the same patient. (a) Pneumothorax with a chest tube (support device), marked by yellow arrows, (b-c) Pneumothorax without a chest tube, (d) neither Pneumothorax nor a chest tube. Pneumothorax regions are outlined in red. \textbf{Visual characteristics unrelated to the pathology or medical device remain largely consistent across the images.}}
\label{figs:chexpert_patient_id}
\end{figure*}

Metadata can provide grouping keys for samples that share visual characteristics. In medical imaging, patient ID is a particularly useful example. Chest X-ray datasets often contain multiple scans of the same patient acquired before and after a medical procedure. These scans may differ primarily in pathology or the presence of medical devices, while other patient-specific visual characteristics remain relatively consistent. This motivates incorporating patient ID into the spurious-correlation mitigation process.
\Cref{figs:chexpert_patient_id} illustrates multiple scans from the same patient.
More broadly, the same principle can be applied beyond medical imaging whenever metadata captures meaningful relationships among visually similar samples, as demonstrated by using bird species in the Waterbirds dataset.

\end{document}